\documentclass[letterpaper, 10 pt, conference]{ieeeconf}  

\IEEEoverridecommandlockouts                              

\usepackage{graphicx} 
\usepackage{amsmath} 
\usepackage{amssymb}  
\usepackage{cite}
\usepackage{subcaption}
\usepackage{hyperref}
\usepackage{caption}

\newcommand{\comment}[1]{}
\usepackage{color}

\newcommand{\mcot}{c_\text{mt}}

\title{\LARGE \bf
Anisotropic body compliance facilitates robotic \\sidewinding in complex environments
}

\author{Velin Kojouharov$^{1,*}$, Tianyu Wang$^{1,*}$, Matthew Fernandez$^{1}$, Jiyeon Maeng$^{1}$, Daniel I. Goldman$^{1}$
\thanks{*These authors contributed equally to this work}
\thanks{$^{1}$All authors are with Georgia Institute of Technology, Atlanta, GA 30332, USA. {\{\tt\small velinkojouharov, tianyuwang, mfernandez64, jmaeng8\}@gatech.edu, daniel.goldman@physics.gatech.edu}}
}

\begin{document}
\maketitle
\thispagestyle{empty}
\pagestyle{empty}

\begin{abstract}

Sidewinding, a locomotion strategy characterized by the coordination of lateral and vertical body undulations, is frequently observed in rattlesnakes and has been successfully reconstructed by limbless robotic systems for effective movement across diverse terrestrial terrains. However, the integration of compliant mechanisms into sidewinding limbless robots remains less explored, posing challenges for navigation in complex, rheologically diverse environments. Inspired by a notable control simplification via mechanical intelligence in lateral undulation, which offloads feedback control to passive body mechanics and interactions with the environment, we present an innovative design of a mechanically intelligent limbless robot for sidewinding. This robot features a decentralized bilateral cable actuation system that resembles organismal muscle actuation mechanisms. We develop a feedforward controller that incorporates programmable body compliance into the sidewinding gait template. Our experimental results highlight the emergence of mechanical intelligence when the robot is equipped with an appropriate level of body compliance. This allows the robot to 1) locomote more energetically efficiently, as evidenced by a reduced cost of transport, and 2) navigate through terrain heterogeneities, all achieved in an open-loop manner, without the need for environmental awareness.
\end{abstract}

\section{Introduction}\label{sec:intro}
Sidewinding serves as the primary locomotion strategy for several desert-dwelling viper species~\cite{mosauer1932adaptive,gray1946mechanism,brain1960observations,gans1972sidewinding}, and for other taxa navigating granular surfaces~\cite{jayne1986kinematics,tingle2020facultatively}.
During sidewinding, snakes generate vertical and lateral undulations in the body, i.e., propagating two waves in the vertical and horizontal planes simultaneously, following a two-wave template~\cite{marvi2014sidewinding,astley2015modulation}.
This coordinated body movement leads to the formation of alternating body lifting and static contact, providing traction on the substrate for robust locomotion.

\begin{figure}[t]
\centering
\includegraphics[width=1\columnwidth]{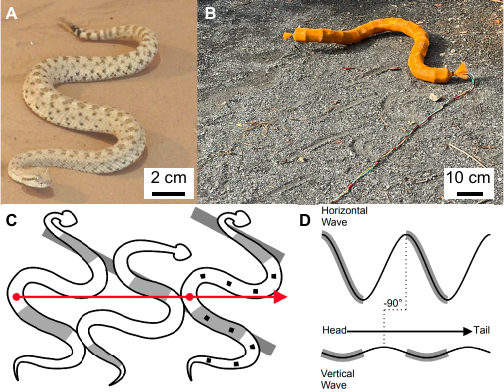}
\caption{Mechanically intelligent limbless robot, inspired by sidewinding snakes, capable of performing sidewinding locomotion in diverse, rheologically complex terrestrial environments. 
(\textbf{A}) The sidewinding behavior observed in rattlesnakes. (\textbf{B}) The sidewinding locomotion of the robot on granular media. (\textbf{C}) A diagram of sidewinding motion. Gray areas in the body indicate static contact with the substrate, and white areas represent body segments lifted and in motion. Gray rectangles denote tracks. The red arrow shows the center of mass direction of motion. Reproduced from~\cite{astley2015modulation}. (\textbf{D}) A diagram of the vertical and horizontal waves propagating from head to tail in sidewinding, characterized by a $\pi/2$ phase difference. Grey areas denote static contact. Reproduced from~\cite{astley2015modulation}.}
\label{fig:snake_and_robot}
\vspace{-1.5em}
\end{figure}

Sidewinding is of great interest for limbless robots (snake robots) to replicate~\cite{burdick1994sidewinding,lipkin2007differentiable,ariizumi2017dynamic,rozaidi2023hissbot}. 
Unlike lateral undulation, which requires drag anisotropy to generate thrust, typically achieved with wheels~\cite{ma2001development,crespi2008online} or scales~\cite{serrano2015incorporating}, sidewinding is a form of locomotion capable of producing translational movement under isotropic friction condition. 
In sidewinding, instead of maintaining consistent body contact with the substrate as in lateral undulation, adjusting the coordination between vertical and horizontal waves enables the body to establish and break contact with the substrate. 
This feature facilitates the design and planning of contact patterns for effective and robust locomotion~\cite{chong2021frequency,chong2021moving}. 
However, research on robotic sidewinding has predominantly focused on homogeneous substrates, while negotiating obstacles during sidewinding remains less explored and challenging.

As compliant body behaviors have been observed in sidewinder rattlesnakes during obstacle negotiation in previous research, it is hypothesized that, in addition to the modulation of gait parameters, robotic sidewinders require body compliance to navigate obstacle-rich environments~\cite{astley2020side}. Previously, a serially linked (joint actuated) limbless robot was used to model the sidewinding rattlesnakes with the method of amplitude modulation~\cite{astley2020side} but ultimately failed to replicate the compliant behaviors exhibited by the snakes when faced with an array of obstacles, due to the lack of sensing capability. This further motivated the idea that compliance is key to sidewinding through obstacle-rich terrains. 

While compliant sidewinding remains less explored, one major approach to achieving compliant lateral undulatory locomotion in limbless robots is through ``computational intelligence," which involves real-time tuning of the body shape in response to obstacles based on proprioceptive sensory feedback (e.g., vision~\cite{xinyu2003control,sartoretti2021autonomous}, contact sensing~\cite{liljeback2014mamba,ramesh2022sensnake}, and joint torque sensing~\cite{travers2018shape,wang2020directional}).  
Recent studies have shown that ``physical intelligence" (PI) or ``mechanical intelligence" (MI)~\cite{sitti2021physical} can be another means of achieving compliant limbless locomotion in complex environments. This approach offloads the complexity of computation and control onto passive body mechanics~\cite{fu2020robotic,schiebel2020robophysical,wang2023mechanical}. Specifically, our prior work~\cite{wang2023mechanical} introduced a bilaterally actuated, cable-driven limbless robot inspired by organismal muscle actuation mechanisms. Through a control scheme for programmable body compliance, we showed that MI simplifies locomotion control for lateral undulation in complex terrestrial terrains.

Inspired by the control simplification achieved through MI in lateral undulation, we hypothesize that MI can similarly enhance obstacle navigation in sidewinding. To validate our hypothesis, we devised a novel 3D cable-driven limbless robot for sidewinding and developed a control scheme for variable body compliance. Through robophysical experiments, we compared the robot's sidewinding performance across varying levels of body compliance and observed that MI emerges when the robot is programmed with an appropriate degree of body compliance, facilitating the negotiation of heterogeneities. Further, by measuring the cost of transport, we demonstrated that MI improves sidewinding energy efficiency.

\section{Robot Design and Control}\label{sec:design}

\begin{figure}[t]
\centering
\includegraphics[width=1\columnwidth]{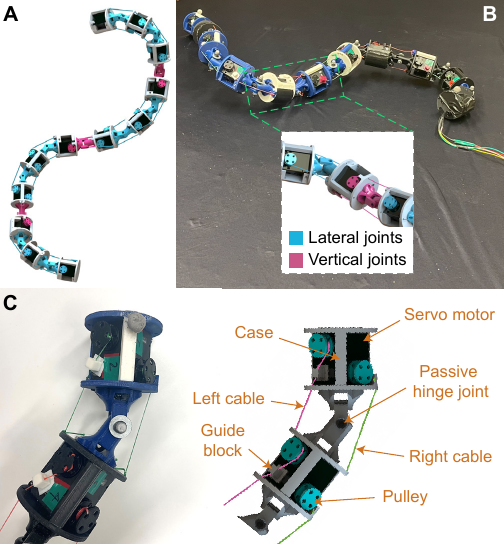}
\caption{Detailed mechanical design of a bilateral cable-driven limbless robot for sidewinding. (\textbf{A}) Computer Aided Design (CAD) representation of the robot. The design features 8 lateral bending joints (cyan) and 3 vertical bending joints (pink) (\textbf{B}) Picture of the robot with zoomed-in view of 2 joints -- one vertical bending and one lateral bending. (\textbf{C}) Picture and labeled schematic of a single robot module. } 
\label{fig:robot_design}
\vspace{-1.5em}
\end{figure}

To test our hypothesis, we designed a modular limbless robot. The robot consists of a series of 12 modules connected by 11 passive hinge joints (total length 1.31 m). There are two types of joints on the robot: vertical bending joints and lateral bending joints, each with one rotational degree of freedom rotation in their respective planes. The combination of these two bending joints allows the robot to simultaneously propagate waves in the horizontal and vertical planes -- necessary to produce a sidewinding gait. The vertical and lateral joints are evenly spaced along the body, where joints 3, 6, and 9 are vertical bending with the remaining 8 being lateral bending (Fig.~\ref{fig:robot_design}A). The higher number of lateral bending joints allows us to achieve much higher curvature in the horizontal plane compared to the vertical plane, similar to what has been observed in sidewinding rattlesnakes~\cite{astley2015modulation}. This gives this robot an advantage in replicating the snake’s gaits compared to previous sidewinding limbless robots that use alternating vertical and lateral bending modules~\cite{marvi2014sidewinding,chong2021frequency}.

\subsection{Module Components} 
Aside from the orientation of the joints (vertical vs. lateral bending), all modules are built identically (length of 10 cm and diameter of 7.5 cm). Each module has a 3D-printed PLA case that houses one DYNAMIXEL 2XL430-W250-T (ROBOTIS), which packages 2 independently controlled servo motors. Each servo motor has a pulley (9.5 mm inner diameter) that is spooled with a non-elastic fishing line (Rikimura) which has negligible shape memory and deformation response to stretching. The other end of each of the two lines is attached to the following case. 

\subsection{Bilateral Cable Actuation}

A majority of existing limbless robots employ a ``joint actuation"  mechanism which actuates each joint in the spine with a rotary motor~\cite{hirose2004biologically,wright2007design,transeth2008snake,takemori2022adaptive}. Alternatively, bilateral cable actuation has recently been used in the design of limbless robots as a way to introduce compliance~\cite{schiebel2020robophysical,wang2023mechanical}. The sidewinding robot presented in this work features a decentralized bilateral actuation mechanism, i.e., each joint is actuated with two independently controlled cables. Thus, thee robot moves through coordinating the shortening and lengthening of each cable.

\subsection{Power and Communication}
The robot is powered by a DC power supply with 11.1 V and receives control signals transmitted from a PC via U2D2 (ROBOTIS). Each servo motor is connected in series with internal wiring running through the joints, resulting in minimal electrical harnessing along the robot's body. The power and communication lines are tied together to create the tether for the robot. 

\subsection{Sidewinding Gait Template}

To implement a sidewinding gait on our robot, we used a two-wave template that is widely used in sidewinding robots~\cite{astley2015modulation,marvi2014sidewinding,astley2020side},
\begin{equation}
\begin{aligned}
    \alpha_H(i,t) &= A_H\sin\left(2\pi\xi_H \frac{i}{N_H}-2\pi\omega t\right), \\
    \alpha_V(i,t) &= A_V\sin\left(2\pi\xi_V \frac{i}{N_V}-2\pi\omega t - \frac{\pi}{2}\right),
\end{aligned}
\label{eq:template}
\end{equation}
where subscripts $H$ and $V$ refer to horizontally and vertically oriented motors, respectively; $\alpha$ represents joint angle; $i$ is joint index; $t$ is time; $A$, $\xi$ and $\omega$ is the amplitude, the spatial and temporal frequencies of the corresponding wave; and $N$ is the total number of joints in the corresponding plane. 

\begin{figure}[t]
\centering
\includegraphics[width=0.8\columnwidth]{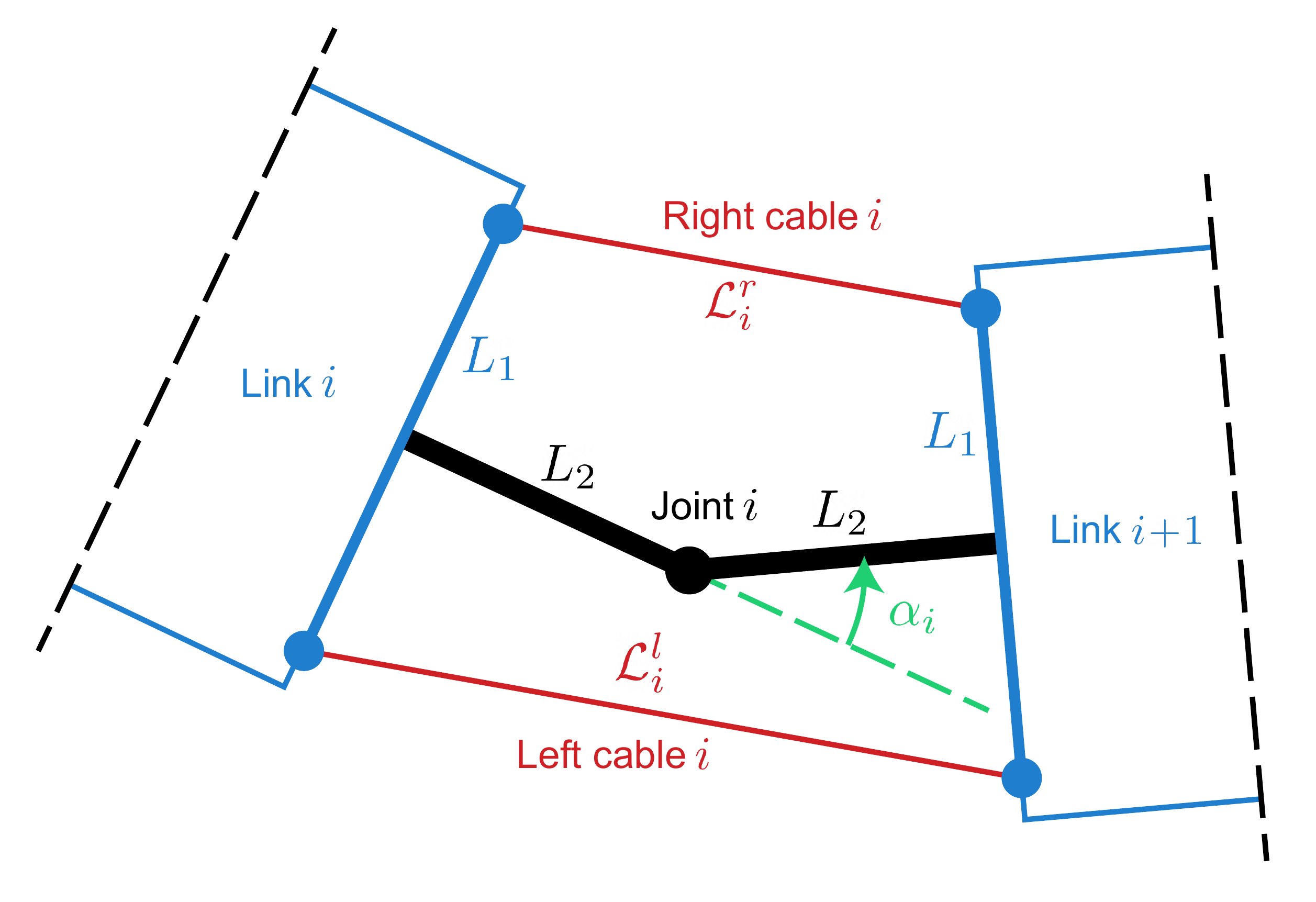}
\caption{Geometry of an individual joint for the calculation of cable lengths to form the suggested joint angle $\alpha$. Reproduced from~\cite{wang2023mechanical}.}
\label{fig:joint_geometry}
\vspace{-2em}
\end{figure}

To accurately form a joint angle $\alpha$ as defined in Eq.~\ref{eq:template}, we need to adjust the lengths of the left and right cables around the joint so that they both are shortened. Since the deformation of cables in the robot is negligible, the lengths of the left cable ($\mathcal{L}^l$) and right cable ($\mathcal{L}^r$) are determined by the robot's geometry as shown in Fig.~\ref{fig:joint_geometry}, following
\begin{equation}
\begin{aligned}
    \mathcal{L}^l(\alpha) &= 2\sqrt{L_{1}^2 + L_{2}^2} \cos\left[-\frac{\alpha}{2}+\tan^{-1}\left(\frac{L_{1}}{L_{2}}\right)\right],\\
    \mathcal{L}^r(\alpha) &= 2\sqrt{L_{1}^2 + L_{2}^2} \cos\left[\frac{\alpha}{2}+\tan^{-1}\left(\frac{L_{1}}{L_{2}}\right)\right].
\end{aligned}
\label{eq:ExactLength}
\end{equation}

\subsection{Programmable Body Compliance}
\begin{figure}[t]
\centering
\includegraphics[width=1\columnwidth]{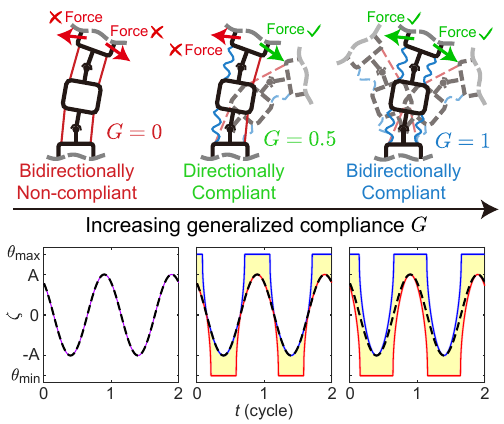}
\caption{Schematic of representative compliant robot states under varied generalized compliance $G$: bidirectionally non-compliant ($G = 0$), where a joint does not admit force in either direction so that the joint angle follows the template trajectory (dashed line); directionally compliant ($G = 0.5$) where a joint only admits force that bends the joint further (to form a larger joint angle as shown by yellow region); and bidirectionally compliant ($G = 1$), where a joint admits force in both directions in an anisotropic manner (to form either a smaller or larger joint angle as shown by yellow region). Reproduced from~\cite{wang2023mechanical}.}
\label{fig:explain_G}
\vspace{-1.5em}
\end{figure}

Based on Eq.~\ref{eq:ExactLength}, we can implement accurate body postures in sidewinding gait on our robot. As mentioned previously, bilateral actuation allows us to program body compliance via coordinately loosening cables. Extending the implementation of the generalized compliance variable ($G$) defined in~\cite{wang2023mechanical} to our sidewinder robot, we can quantify the body compliance in the robot using $G$. The cable length control scheme in this work is then given by
\begin{equation}
\begin{array}{l}
L_{H,i}^l(\alpha_{H,i}) = \left\{\begin{array}{llc}{\mathcal{L}_{H,i}^l(\alpha_{H,i}),} & {\text{if } \alpha_{H,i} \leq -\gamma} \\ {\mathcal{L}_{H,i}^l[-\min(A, \gamma)]}\\{+l_0\cdot[\gamma + \alpha_{H,i}],} & {\text{if } \alpha_{H,i} > -\gamma}\end{array}\right. \\ 
L_{H,i}^r(\alpha_{H,i}) = \left\{\begin{array}{llc}{\mathcal{L}_{H,i}^r(\alpha_{H,i}),} & {\text{if } \alpha_{H,i} \geq \gamma} \\ 
 {\mathcal{L}_{H,i}^r[\min(A, \gamma)]}\\{+l_0\cdot[\gamma - \alpha_{H,i}],} & {\text{if } \alpha_{H,i} < \gamma}\end{array}\right. \\
L_{V,i}^l(\alpha_{V,i}) = \mathcal{L}_{V,i}^l(\alpha_{V,i})\\
L_{V,i}^r(\alpha_{V,i}) = \mathcal{L}_{V,i}^r(\alpha_{V,i})
\end{array}
\label{eq:policy}
\end{equation}
where superscripts $l$ and $r$ refer to left and right, respectively; $\gamma$ is short for $(2G-1)A_H$; and $l_0$ is a design parameter which we fix over this work as 41.8 mm/rad. Following Eq.\ref{eq:policy}, the robot can achieve three representative compliance states with varied $G$ (Fig.~\ref{fig:explain_G}): 1) bidirectionally non-compliant ($G=0$), where each joint angle strictly follows the trajectories suggested by Eq.~\ref{eq:template}; 2) directionally compliant ($G=0.5$), where the joints are only allowed to be perturbed to form a larger angle than suggested; and 3) bidirectionally compliant ($G=1$), where the joints are allowed to be perturbed in both directions, in an anisotropic way regulated by Eq.\ref{eq:policy}. For a detailed discussion of this length control scheme we refer to~\cite{wang2023mechanical}. Note that in this work, we only enabled programmable compliance on the horizontal joints, whereas vertical joints remain non-compliant ($G=0$) for all time.

\section{Results}\label{sec:results}

\subsection{Flat Terrain Experiment}

As hinted in previous work where body compliance can improve lateral undulation locomotion efficiency in diverse environments~\cite{wang2023mechanical}, we started with testing the robot's sidewinding performance on flat terrain with varied generalized compliance value $G$. In this experiment, we fixed the parameters in Eq.~\ref{eq:template} as $A_H = 75^\circ, \xi_H = 1, A_V = 25^\circ, \xi_V = 1$, with which the robot's body shape can approximate that observed from rattlesnakes (video included in the supplementary video)~\cite{astley2015modulation}. We quantify the performance using locomotion speed and mechanical cost of transport, quantities that are commonly used to study both biological and robotic locomotion~\cite{collins2005efficient,seok2013design,saranli2001rhex}. 

\begin{figure}[t]
\centering
\includegraphics[width=1\columnwidth]{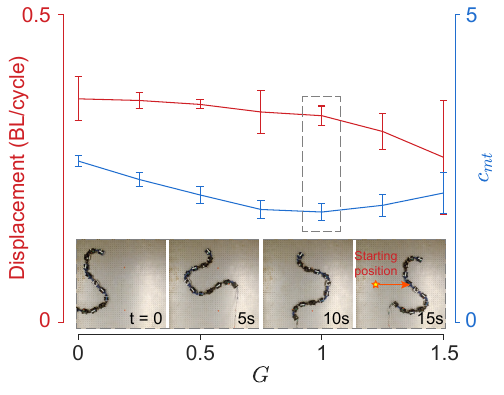}
\caption{Sidewinding locomotion speed (red) and mechanical cost of transport $\mcot$ (blue) as a function of body compliance $G$. Locomotion speed is measured by the averaged center of mass displacement normalized by the body length of the robot over a gait cycle. Mechanical cost of transport is a unit-less quantity calculated by the work done by cables divided by the product of the robot's weight and distance traveled. Error bars represent standard deviations. The inset shows a time lapse of the bilaterally compliant ($G = 1$) robot sidewinding on hard ground.}
\label{fig:disp_and_cot}
\vspace{-1.5em}
\end{figure}

We set up a similar experiment shown in Fig.~\ref{fig:disp_and_cot} by running the sidewinding gait on the robot on a flat surface with Coulomb friction ($\mu\approx 0.7$). We varied the generalized compliance of the robot in the lateral bending joints, from $G = 0$ (fully rigid) to $G = 1.5$ (very compliant) with an increment of 0.25. We ran three trials for each $G$ value and in each trial the robot sidewinds two gait cycles. We attached 13 markers evenly on the robot's body and recorded the robot's motion using OptiTrack motion tracking system. We then averaged each marker's displacement to calculate the robot's center of geometry displacement. To calculate mechanical cost of transport ($\mcot$), we used the equation $\mcot = W/mgd$, where $W$ is the work done by cables which is estimated using the torque sensor reading from the servo motor, $mg$ is the robot's weight, and $d$ is the displacement.

We found that unlike in lateral undulation, when sidewinding in an open environment, having compliance in the body can decrease the mechanical cost of transport in open, hard-ground environments. While the fully rigid body results in a slightly higher displacement (0.476 m/cycle) compared to the $G = 1$ robot (0.400 m/cycle), the work done by the pulleys in the $G = 1$ is far less, resulting in a consistent decrease in the mechanical cost of transport as generalized compliance increases. The value of $G = 1$ was the local minima of the cost of transport. After $G = 1$, the robot can no longer maintain the desired contact pattern for effective sidewinding, resulting in much lower displacements per body cycle (for $G = 1.5$, the robot only translates 0.351 m/cycle). This result gave us the basis for selecting what generalized compliance parameters to use in later experiments. Given that sidewinding efficiency tends to break down after $G = 1$, for the following experiments, we will be comparing three G values: 0, 0.5, and 1.

\subsection{Obstacle Terrain Experiment Setup}

To verify our hypothesis that mechanical intelligence induced by the body compliance can enhance obstacle navigation in sidewinding, we set up a model heterogeneous environment for the robot: a level pegboard base ($L=2.4$ m, $W=1.2$ m) with a row of obstacles (5 cm diameter PVC pipes) as depicted in Fig.~\ref{fig:peg_variation}A. In this series of experiments, we fixed the parameters in Eq.~\ref{eq:template} as $A_H = 75^\circ, \xi_H = 1, A_V = 25^\circ, \xi_V = 1$. The parameters were selected so that the ratio of the wavelength displayed in robot and the obstacle spacing roughly matches with that observed from rattlesnakes ($\sim$0.8, video included in the supplementary video)~\cite{astley2015modulation}. Further, the robot is wrapped with a mesh skin (4 cm ID expandable sleeving, McMaster-Carr) to create a smoother contact surface between the robot and the environment. 

\subsection{Experiment with Varied Obstacle Spacing}

A total of 15 sets of trials were conducted \textendash{} 5 different obstacle spacings (60, 65, 70, 75, and 80 cm) each with 3 different generalized compliance values ($G = 0, G = 0.5, G = 1$). Given that the attack angle and initial condition of the body could affect the ability of the robot to traverse through the obstacles, we selected five different initial positions and orientations to start the gait for each set of trials. For our experiments, the criterion for success was to have the entire body clear the line connecting centers of obstacles. If the robot fails to clear the center line of the obstacles after 10 gait cycles or if the robot became jammed between two obstacles, the experiment was classified as a failure. In every set of trials, the traverse probability represents the percentage of successful outcomes out of five initial positions.

\begin{figure*}[t]
\centering
\includegraphics[width=1\textwidth]{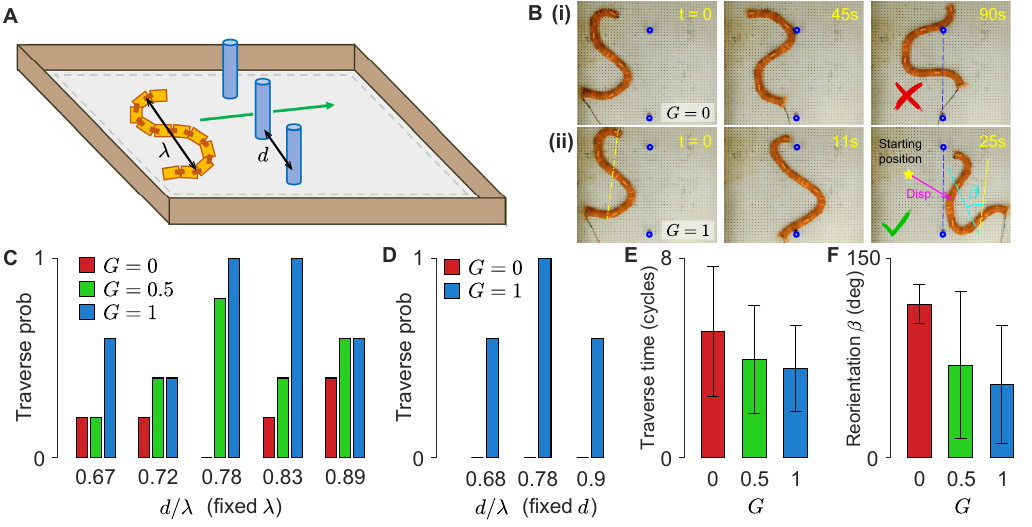}
\caption{Robot performance when sidewinding through an array of obstacles. (\textbf{A}) Diagram of the experimental setup. Obstacle spacing $d$, robot initial condition, robot wavelength $\lambda$ and the generalized compliance parameter $G$ were varied for different experiments. (\textbf{B}) Time-lapse photos of (i) a failure ($G=0$) and (ii) a success ($G=1$). Success is the entire robot body passing the center line intersecting the obstacles. (\textbf{C}) The traverse (success) probability of the robot for different ($G$) values across different obstacle spacing (normalized by the robot's wavelength). (\textbf{D}) The traverse (success) probability of the robot for different ($G$) values with different robot wavelengths and fixed obstacle spacing of 70cm (the axis is obstacle spacing normalized by the robot's wavelength). We tested three different gaits with $A_H=82.5,75,67.5$ deg and $\xi_H=1.1,1.0,0.9$, respectively, which are noted by their corresponding wavelengths of the robot body shape $\lambda=79, 91, 104$ cm. (\textbf{E}) The average traverse time (in number of cycles) to traverse through the obstacles for each successful trial, sorted by $G$ value. (\textbf{F}) The average robot reorientation angle (in degrees) for each successful trial, sorted by $G$ value.}
\label{fig:peg_variation}
\vspace{-1.5em}
\end{figure*}

Our experiment results indicate that, across different obstacle spacings, having a more compliant body led to a higher traverse probability Fig.~\ref{fig:peg_variation}C. Moreover, the robot that has anisotropic bidirectional compliance outperforms others because it allows body joints to comply with the obstacles in different directions. We observed that in the bidirectionally compliant robot 1) the interactions with the obstacles led to less drastic deviations from the robots initial trajectory, and 2) the body compliance allowed the robot to deform its body to squeeze through obstacles that are tighter than the robot's body length before deformation. Contrary, the two primary failure modes that were observed with the non-compliant robot were: 1) the robot was not able to deform its body enough to squeeze between two obstacles or 2) because the robot cannot absorb the impact of obstacle collisions, it rapidly reorients its body into an undesirable position, causing it to jam. Both of the failure modes are mitigated by increased compliance. Fig.~\ref{fig:peg_variation}F shows the average reorientation angle in successful trials for different $G$ parameters of the robot. With $G = 0$, the average reorientation angle was $115.5 \pm 14.6$ degrees, with $G = 0.5$ it was $69.6 \pm 55.2$ degrees, and with $G = 1$ it was $55.1 \pm 44.5$ degrees. The average reorientation angle was lower for the more compliant robot because it locally deforms its bodies to mitigate harsh obstacle contacts that cause reorientation. Further, across all trials, the robot with bilateral compliance ($G = 1$) had lower a average number of cycles to traverse ($3.59 \pm 1.73$ cycles to traverse, in the success trials) compared to both the directionally compliant ($3.96 \pm 2.17$ cycles to traverse, in successful trials) and the non-compliant robot ($5.07 \pm 2.61$ cycles to traverse, in the success trials) as shown in Fig.~\ref{fig:peg_variation}E. Overall, increased body compliance helps to prevent and mitigate reorientation due to obstacle interaction and decreases the number of cycles necessary for the robot to traverse through the obstacle array.

Note that while body compliance shows its advantages across experiments with varied obstacle spacings, by far the highest traverse probability for the robot was at 70 and 75 cm obstacle spacing, the same obstacle spacing ratio as what was observed in the biological experiments. We hypothesize that having compliance alone is not exclusively sufficient for obstacle-rich environments when sidewinding. Instead, choosing the ``appropriate" gait parameters based on the heterogeneities present in the environment is also important. On the other hand, appropriate gait parameters alone cannot guarantee traversal, as the traverse probability for the $G = 0$ trials consistently remained below 20\%. Thus, our results indicate that in order to achieve effective locomotion within complex environments, a sidewinding robot needs the synergy of computational intelligence (to select appropriate parameters) and mechanical intelligence (for passive body mechanics and compliant body-environment interactions). 

\subsection{Experiment with Varied Gait Parameters}
To further validate that the effect of body compliance is not exclusive to specific gait parameter choices, we varied the spatial frequency and amplitude of the horizontal wave and ran experiments at the 70 cm obstacle spacing. Without the loss of generality, we chose $A_H = 82.5^\circ, 75^\circ, 67.5^\circ$ and $\xi_H = 1.1, 1, 0.9$, respectively, while $A_V$ and $\xi_V$ remained unchanged. As in the previous tests, each of these experiments was repeated with 5 different initial conditions, and we compared the robot's performance with no compliance ($G = 0$) and with anisotropic bidirectional compliance ($G=1$). 

Remarkably, the bidirectionally compliant robot produced traverse probabilities larger than 60\% for all parameter combinations as shown in Fig.~\ref{fig:peg_variation}D. While for all three gait parameter combinations, the non-compliant robot failed to get through in every trial. This result suggests that with an appropriate level of body compliance $G$, robot performance can remain robust for an increased range of parameters. Even without an ``optimal" choice in gait parameters for a particular environment, body compliance can help facilitate effective locomotion.

\subsection{Natural Terrain Experiment}
Lastly, we conducted a series of open-loop outdoor experiments to examine the potential applications of sidewinding with anisotropic bidirectional compliance in complex natural terrains. We tested the robot in two different terrains: 1) pine straw with small ferns and 2) coarse granular media.
These environments imitate what the robot could encounter during future applications such as planetary exploration, environmental monitoring, and open-field search-and-rescue tasks. Each of the trials was performed with bidirectional compliance ($G = 1$) in the horizontal bending joints and non-compliant vertical bending joints. Similar to the observations in indoor experiments, bidirectional compliance allowed for effective negotiation of irregularities, as the robot body is more likely to deform and deflect from the harsh contact with surrounding obstacles. Our outdoor experiments demonstrated the robot's locomotion capability and potential for practical applications.

\begin{figure}[t]
\centering
\includegraphics[width=1\columnwidth]{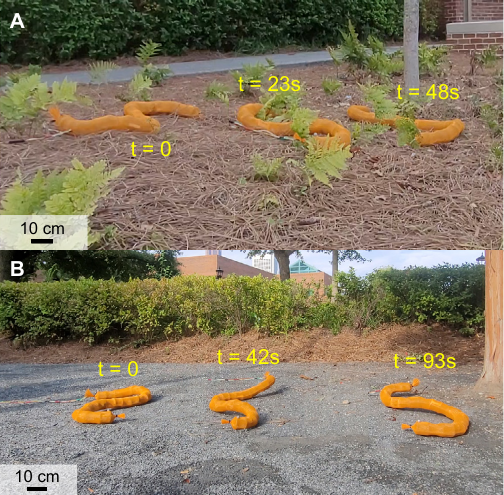}
\caption{The robot demonstrates its capability of sidewinding in complex natural environments with bidirectional compliance ($G = 1$). (\textbf{A}) Time lapsed images of the robot traversing pine straw and fern environment. (\textbf{B}) Time lapsed images of the robot traversing coarse granular media environment. }
\label{fig:outdoor_demo}
\vspace{-1.5em}
\end{figure}


\section{Discussion and Conclusion}\label{sec:conclusion}

Obstacle negotiation in complex, natural environments remains challenging for limbless robots. Prior research on limbless robot locomotion has attempted to tackle these challenges through a variety of methods, relying on online gait parameter tuning based on precise real-time proprioceptive sensory feedback of the environment~\cite{xinyu2003control,sartoretti2021autonomous}. These methods often require high onboard computational capabilities or sufficient prior knowledge of the environment for effective locomotion. Contrary, recent work in lateral undulation has focused on offloading the computational complexity needed for obstacle negotiation to mechanically intelligent, compliant robot design~\cite{wang2023mechanical}. 

In this work, we focused on introducing compliance to sidewinding to simplify the control needed in complex terrains. By incorporating compliance into the robot, we simplify the control process, enabling the robot to sidewind effectively with open-loop controls over a range of heterogeneities in the environment. Our approach utilizes a traveling wave template for both vertical and horizontal waves that exhibits low sensitivity to variations in wave parameters. We observed that across the various robot sidewinding experiments, by introducing compliance we achieve both more energetically efficient locomotion on hard ground, and improved navigation through heterogeneities in both lab and outdoor terrains. We hypothesize that when sidewinding obstacle-rich environments, having compliance in the lateral wave helps minimize the effect of harsh robot-environment interactions, allowing the robot to either 1) squeeze through obstacles or 2) brush by them without having large changes in body orientation. The robot's ability to exploit its compliance to improve open-loop sidewinding performance across these various terrains makes it mechanically intelligent.  

Notice that in this work, the generalized compliance parameter ($G$) was only varied in the lateral joints, not the vertical joints. Sidewinding requires careful coordination of horizontal and vertical waves along the body to establish and break contact with the substrate. Implementing the same compliance strategy in the vertical direction as the one used in the lateral joints negatively affected the robot's ability to sidewind. We hypothesize that this is because the contact pattern determined by the suggested gait is disturbed by unwanted ground contact brought by vertical compliance. Instead of remaining above the ground, vertical bending joints ``droop down". However, we assume there could be better compliant strategies for vertical waves during sidewinding so that the contact pattern can be preserved while the energy consumption can go down.

This work also builds a strong framework for designing multi-modal compliant limbless robots capable of multiple modes of limbless locomotion (e.g., sidewinding, lateral undulation, etc.). Previous work has shown that compliance can improve obstacle negotiation in highly obstacle-dense terrains when using lateral undulation~\cite{wang2023mechanical}. This work suggests that the same bilateral actuation strategy can be used to aid sidewinding in both open and obstacle-rich environments. By designing a robot capable of exploiting body compliance to be mechanically intelligent in both sidewinding and lateral undulation, we can get closer to creating agile, robust, and capable limbless robots for real-world applications such as search-and-rescue, planetary exploration, and inspection. 

More generally, modeling mechanics and interactions involved in biological limbless locomotion are challenging, making limbless robots good tools (as ``robophysical" models) for revealing fundamental principles underlying limbless locomotion~\cite{aguilar2016review,marvi2014sidewinding,wang2020omega,wang2022generalized,wang2023mechanical}. To this end, this robot has the potential to serve as a model to study snake sidewinding. With a bilaterally cable-driven robot we can systematically test locomotor performance with varied gait parameters and level of body compliance, which is impossible to carry out with animals. Through comparison across robotic and biological systems, this robot can help us learn sidewinding snakes' kinematics, dynamics, and even physiology, deepening our understanding of their locomotion in complex terrains. 


\bibliographystyle{IEEEtran}

\bibliography{ICRA2024_CompliantSidewinding}

\end{document}